\documentclass[10pt,twocolumn,letterpaper]{article}

\usepackage{wacv}
\usepackage[table]{xcolor}
\usepackage{times}
\usepackage{epsfig}
\usepackage{graphicx}
\usepackage{amsmath}
\usepackage{amssymb}
\usepackage{multirow}
\usepackage{adjustbox}
\usepackage{booktabs}
\usepackage{tabularx}
\usepackage{enumitem}
\usepackage{multicol}
\usepackage{subcaption}
\makeatletter
\@namedef{ver@everyshi.sty}{}
\makeatother
\usepackage{tikz}
\usepackage[linesnumbered,ruled,vlined]{algorithm2e} 
\DontPrintSemicolon 
\usepackage[nocompress]{cite} 
\usepackage{nicematrix}
\usepackage{authblk}
\usepackage[accsupp]{axessibility}
\usepackage[title]{appendix}

\def\wacvPaperID{27} 

\wacvalgorithmstrack   

\wacvfinalcopy 

\ifwacvfinal
\usepackage[breaklinks=true,bookmarks=false]{hyperref}
\else
\usepackage[pagebackref=true,breaklinks=true,colorlinks,bookmarks=false]{hyperref}
\fi

\begin{document}

\title{Neural Weight Search for Scalable Task Incremental Learning}

\author{Jian~Jiang}
\author{Oya~Celiktutan}
\affil{Department of Engineering, King's College London, London, UK \authorcr
  \{\tt jian.jiang, oya.celiktutan\}@kcl.ac.uk}

\maketitle
\thispagestyle{empty}

\begin{abstract}
Task incremental learning aims to enable a system to maintain its performance on previously learned tasks while learning new tasks, solving the problem of catastrophic forgetting. One promising approach is to build an individual network or sub-network for future tasks. However, this leads to an ever-growing memory due to saving extra weights for new tasks and how to address this issue has remained an open problem in task incremental learning. In this paper, we introduce a novel Neural Weight Search technique that designs a fixed search space where the optimal combinations of frozen weights can be searched to build new models for novel tasks in an end-to-end manner, resulting in a scalable and controllable memory growth. Extensive experiments on two benchmarks, i.e., Split-CIFAR-100 and CUB-to-Sketches, show our method achieves state-of-the-art performance with respect to both average inference accuracy and total memory cost.~\footnote{Implementation: \href{https://github.com/JianJiangKCL/NeuralWeightSearch}{https://github.com/JianJiangKCL/NeuralWeightSearch}}
\end{abstract}

\section{Introduction}
The last decade has demonstrated the power of deep learning approaches, achieving superior performance in many machine vision tasks. However, modern machine learning algorithms for robots assume that all the data is available during the training phase. On the other hand, the real world is highly varied, dynamic, and unpredictable. It is infeasible to collect enough amount of data to represent all the aspects of the real world. Therefore, robots must learn from their interactions with the real world continuously. Motivated by this, incremental learning (also known as lifelong learning and continual learning) is an emerging research area, which aims to design systems that can gradually extend their acquired knowledge over time through learning from the infinite stream of data~\cite{CL_IBDRR,CL_LUCIR,CL_LwF,CL_CURL,SI2017,CL_AAN}.   

Incremental learning remains a challenging open problem because it demands minimal performance loss for old tasks and a minimal increase in model storage. In other words, the models should be able to adapt to novel tasks efficiently and effectively, while not significantly underperforming on the previously learned tasks, which is known as the problem of catastrophic forgetting. 
\begin{figure}[t]
\centering
\includegraphics[width=0.4\textwidth]{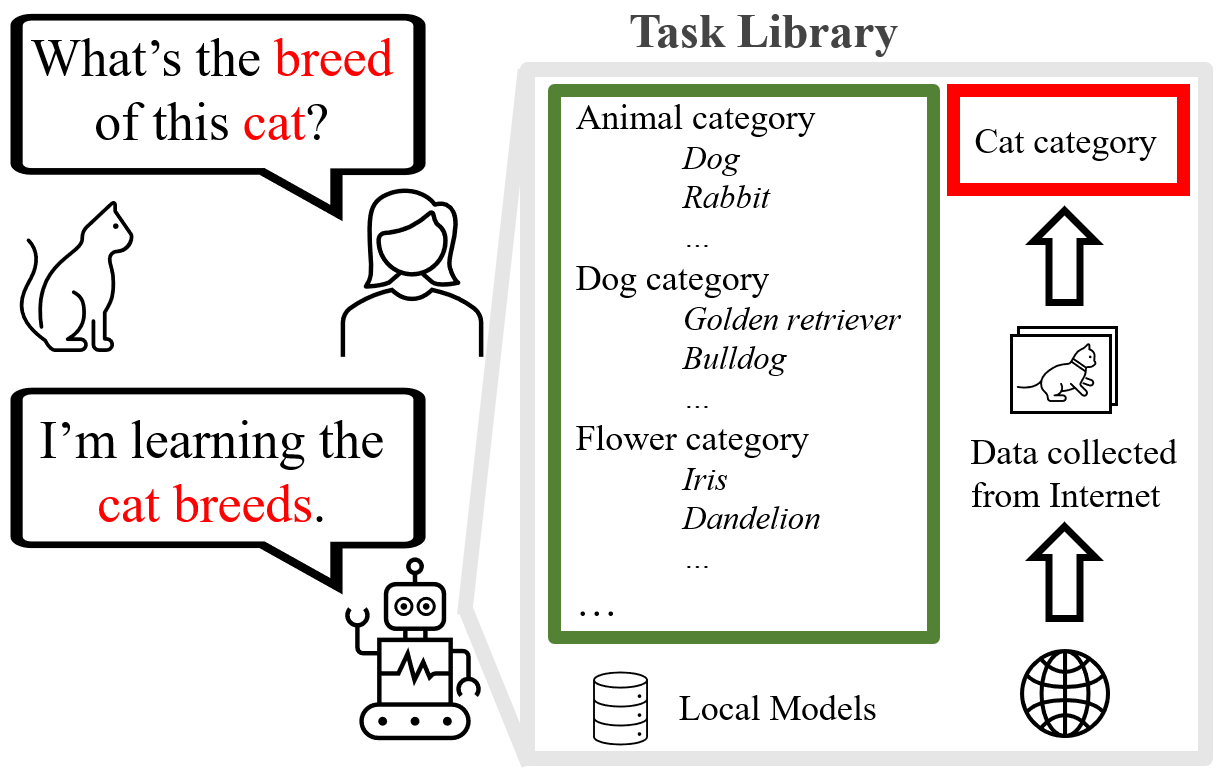}
\caption 
{Envisioned one practical application of the task incremental learning in real-world human-robot interaction settings. To enable such practical applications, this paper aims to address the problem of memory growth. 
}
\vspace{-0.5cm}
\label{fig:interaction}
\end{figure}

There is a significant body of work on incremental learning~\cite{CL_Review_Rahaf,CL_Review_Parisi}. These methods can be divided into two widely used categories based on the learning scenario, namely, class-incremental learning (CIL) and task-incremental learning (TIL). Generally, CIL methods build on a single network model that needs to sequentially learn a series of tasks. One advantage is that they do not require the task id during inference, unlike TIL methods. However, the single shared model will inevitably cause degradation in performance for previously learned tasks. Moreover, state-of-the-art CIL methods, namely, replay-based methods~\cite{CL_AAN,CL_Mnemonics}, require extra memory for saving exemplars per task, e.g., typically a task has $5$ classes and 20 images are saved per class, which will lead to a significant increase in memory as the number of tasks becomes larger. 

TIL methods, on the other hand, learn a separate model for each new task, inherently addressing the problem of forgetting. One downside is that task id is needed during inference. A solution to this is illustrated in Figure \ref{fig:interaction}, where a robot can interactively query its user to determine the task id in both training and inference, which brings about practical applications in human-robot interaction~\cite{CL_application1}. However, building an individual network or sub-network for new tasks leads to an ever-growing memory due to saving extra weights and how to address this issue lies at the heart of TIL research, especially, considering the limitations of robotics platforms, i.e., space constraints.

\begin{figure*}[htb]
\centering
\includegraphics[width=\textwidth]{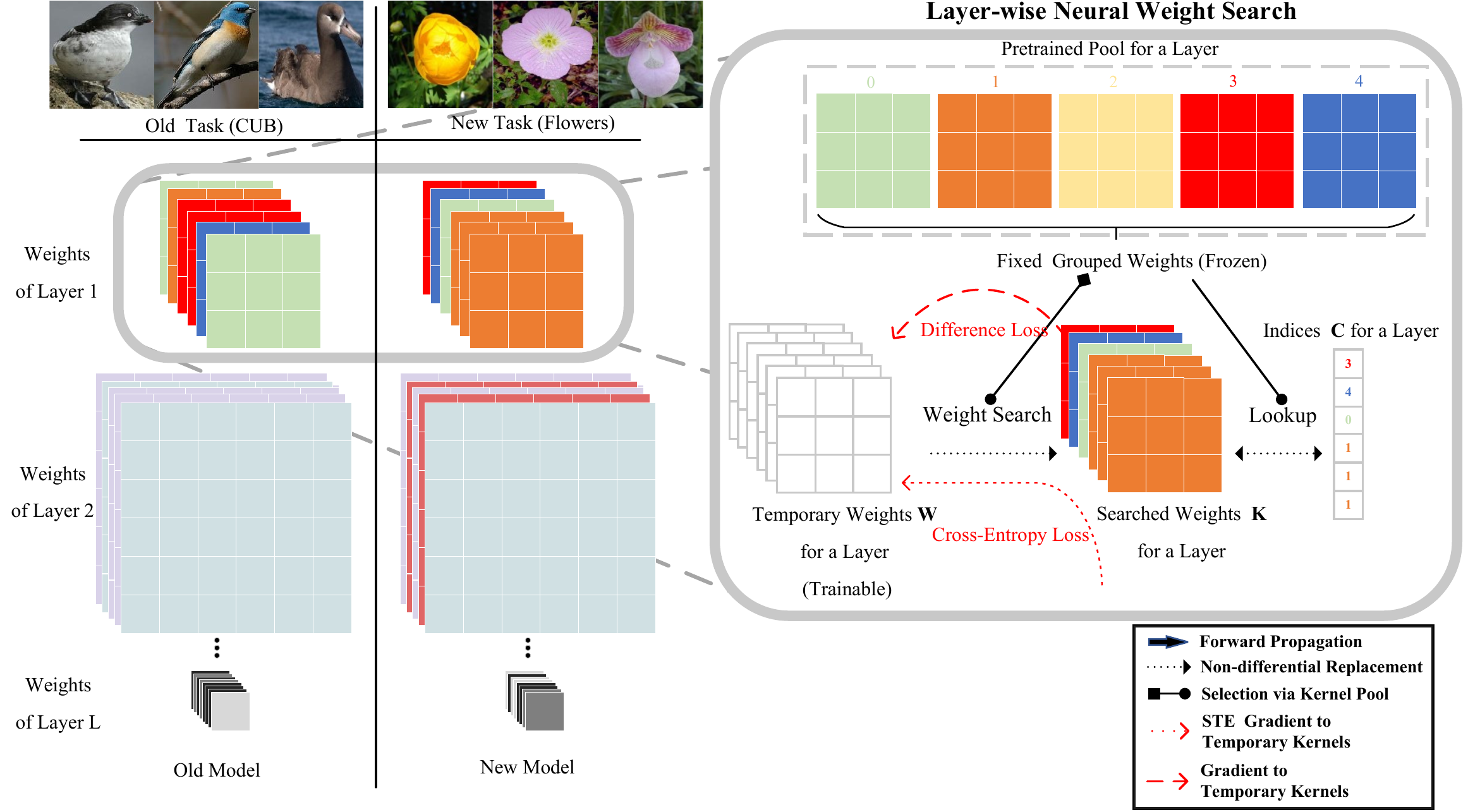}
\caption 
{The illustration of the Neural Weight Search algorithm. In both training and inference, only searched kernel weights are used to process inputs. The temporary kernel weights can be regarded as a scaffold for building a neural network model, which can be discarded after training.}
\label{fig:KP}
\vspace{-0.5cm}
\end{figure*}
Due to its relevance to human-robot interaction, in this paper, we focus on task incremental learning setting, with the goal of alleviating the memory growth problem as a robot encounters new tasks sequentially. To this end, a line of research has focused on dynamically expanding the network for task incremental learning by creating a new model or a subnetwork to learn new information while allowing old models to maintain their performance. Among these methods, Progressive Neural Network (PNN)~\cite{CL_PNN2016} freezes the previously learned models and continuously adds new models for new tasks. One downside of PNN is that it results in substantial growth of parameters with the increasing number of tasks. Recent works~\cite{CL_PiggyBack,CL_PackNet,CL_KSM_yang2021} have focused on learning task-specific kernel masks to transfer and adapt the backbone model to a certain task. PiggyBack~\cite{CL_PiggyBack} aims to learn binary element-wise kernel masks for new tasks. Though, there are two main limitations to this approach. First, binary values might limit the representation capability. Second, the fixed backbone hampers the learning of new tasks. PackNet~\cite{CL_PackNet} aims to solve this problem by freeing up some of the existing weights, which are unused by previous tasks, for learning new tasks. However, the backbone model of PackNet can also run out of learnable weights and might become a `fixed' backbone eventually. To address this issue, Compacting, Picking, and Growing (CPG)~\cite{CL_CPG_hung2019}, grows and prunes the backbone of PiggyBack for incremental learning, which however has a tideous iterative training process and may lead to a significant increase in the storage required for saving models. Taken together, scalable task incremental learning has remained an open problem, particularly, how to utilise previously learned knowledge for new tasks and grow the models in a controllable manner. 

This work approaches this open problem by tapping into a scalable technique, called Neural Weight Search (NWS), which significantly alleviates the memory growth problem while achieving state-of-the-art performance. Differently from the existing methods~\cite{CL_PackNet,CL_PiggyBack,CL_KSM_yang2021}, where a backbone model is used, NWS discards the backbone but maintains frozen layer-wise pools of grouped weights. For each new task, it builds a new model by searching the pools for the optimal combinations of grouped weights. This search is efficiently conducted with the help of a set of temporary kernel weights that are not used for processing inputs and are discarded after the training. Figure~\ref{fig:KP} illustrates the workflow of our proposed approach.

In summary, our main contributions are as follows: (i) We propose a new problem setting named \textit{Neural Weight Search} (NWS). Analogously to Neural Architecture Search (NAS), NWS \textit{automatically searches pretrained weights to build a network}. (ii) The backbone-free design is novel in task incremental learning. Unlike the conventional kernel mask-based methods, where the backbone model is fixed or partially fixed, our method discards the backbone but maintains layer-wise pools of kernel weights, allowing more representation capacity. (iii) NWS enables the re-utilization of weights when building models and achieves scalable performance. Compared to the state-of-the-art method~\cite{CL_KSM_yang2021}, our memory gain can reach up to $82\%$ (including the memory required for saving the pools) and achieves better average accuracy by a margin of $1.9\%$ on two existing benchmarks.

\section{Related Work}
\label{sec:Related_work}
There is a significant body of work on incremental learning, which can be divided into three broad categories, namely, regularization-based (e.g.,~\cite{EWC,SI2017}), architecture-based (e.g.,~\cite{AE_CL,GWR,CL_PiggyBack,CL_PackNet,CL_KSM_yang2021,CL_CPG_hung2019}), and replay-based (e.g.,~\cite{CL_IBDRR,CL_ER,CL_SER,GEM,AGEM,CL_Mnemonics,CL_GFR}). Our proposed approach is at the intersection of two categories, namely,~\textit{regularization-based}, and~\textit{architecture-based} methods. As mentioned before, there are two widely used settings in incremental learning, namely, class-incremental learning (CIL)~\cite{CL_IBDRR,CL_ER,CL_SER,GEM,AGEM,CL_Mnemonics,CL_GFR}, task-incremental learning (TIL)~\cite{CL_PiggyBack,CL_PackNet,CL_KSM_yang2021,CL_CPG_hung2019}. 
In this work, we focus on the task incremental learning setting. 
TIL approaches are generally architecture-based methods (also known as parameter-isolation methods). These methods learn an individual model or a partially original model (e.g., shared low-level layers and individual high-level layers) for each task. Overall, these methods suffer from an uncontrollable growth of the memory when new models are saved. For example, Rusu~\etal proposed Progressive Neural Network (PNN)~\cite{CL_PNN2016} to continuously expand the network by generating a new network for each task while fixing the previously learned networks, which resulted in an uncontrolled growth in parameters and hence poor scalability. 

Recent methods like Piggyback~\cite{CL_PiggyBack}, PackNet~\cite{CL_PackNet}, CPG~\cite{CL_CPG_hung2019}, and KSM~\cite{CL_KSM_yang2021} have aimed to alleviate this problem by introducing learnable masks with a single backbone model. The weights of a new model are generated by multiplying the masks with corresponding weights in the backbone model. PiggyBack~\cite{CL_PiggyBack} fixes the backbone network and learns binary element-wise masks for kernels. First, real-valued masks are generated, which have the size same as kernels. Then a predefined threshold is applied to obtain binary masks. Such masks, namely `hard masks', also result in poor scalability as kernels that can be learned for new tasks are constrained by the fixed backbone model. PackNet~\cite{CL_PackNet}, building upon PiggyBack, uses a strategy to prune weights that are not used by old tasks, allowing to free up those parameters for learning future tasks. However, PackNet does not add more kernels and it reaches a bottleneck when no more parameters are left to be released. The ability of incremental learning is the same as PiggyBack, when it runs out of learnable weights. Hence, PackNet advances PiggyBack but it is still not scalable in the long term. 

Inspired by the aforementioned methods, CPG~\cite{CL_CPG_hung2019} adopts the structure of Piggyback but it enables adaption of the network by iteratively introducing more kernels for new coming tasks and pruning the learned model. The iterative expanding and pruning operations continue until the model reaches a pre-defined inference performance for a task. For example, the pre-defined performance can be defined as the inference accuracy of a set of baseline models individually fine-tuned for the corresponding task. However, CPG has two main limitations. First, prior knowledge about baseline performance is not usually available beforehand. Second, it has a tedious and demanding iterative training process. A recent method called KSM~\cite{CL_KSM_yang2021} uses the setting the same as Piggyback but utilises soft kernel-wise masks that combine binary and real values instead of using binary element-wise kernel masks. Compared to Piggyback, soft kernel-wise masks enhance the incremental learning ability, allowing kernels to adapt to different tasks using richer representations. KSM reaches the state-of-the-art performance on the Split-CIFAR-100~\cite{Cifar100} benchmark and CUB-to-Sketches benchmark~\cite{DS_CUB200,DS_STANFORD_CARS196,DS_OXFORD_FLOWERS102,DS_WIKIART195,DS_SKETCH250,DS_FOOD101}. However, KSM also relies on a fixed backbone model that limits the learning of representations. Unlike the aforementioned kernel-mask-based methods, where the backbone model is fixed or partially fixed, our method discards the use of the backbone. We maintain frozen pretrained kernel weights that are saved in the form of layer-wise pools and these weights can be efficiently reused in various combinations and orderings for different tasks, maximising the plasticity and significantly reducing memory growth.

\section{Problem Definition}
\subsection{Task Incremental Learning (TIL)}
In a general TIL setting, there are $T$ incremental phases where $t_{th}$ phase introduces a new $v_{t}$-way classification task with training data $\mathbf{x}_t$ and labels $\mathbf{y}_t$, where $y_{t} \in \{0, 1, 2, ..., v_{t} - 1\}$. For each task, we aim to learn an individual model $f_{t}$ parameterized by learnanble weights $\boldsymbol{\theta}_{t}$ with the objective, 
$
\arg \min_{\boldsymbol{\theta}_{t}} \mathcal{L}(f_{t}(\mathbf{x}_{t}), \mathbf{y}_{t}),
$
where $\mathcal{L}$ is a classification loss. During the inference, each task-specific model $f_{t}$ is evaluated on the corresponding test data set $\mathbf{x}^{test}_t$ and $\mathbf{y}^{test}_t$.

\subsection{Neural Weight Search}
\label{subsec:nws}
To address the limitations of existing approaches (see Section~\ref{sec:Related_work}), we define a new problem setting named Neural Weight Search (NWS) for TIL. Instead of relying on a fixed or partially fixed backbone and learning a mask for each new task, NWS aims to build a new model by searching for an optimal combination of weights from fixed and stored weights pretrained on a large-scale image dataset. These weights can be reused (without updating the values of weights) for any upcoming new task. Treating each weight scalar as an element to search for would lead to significantly large search space. In practice, we search for grouped weights and a group of weights can either be a convolution kernel, a filter or even a layer. Briefly, we first design a fixed search space of indexed grouped neural weights. Then, we search for an optimal combination of grouped weights in the search space to build a new model for a new task, where the same grouped weights can be shared within the same task as well as across different tasks. 

As illustrated in Figure~\ref{fig:KP}, for each layer $l$, we hold a search space of $n^l$ indexed groups of weights $\mathbf{M}^{l} =\{\mathbf{m}_{1}, \mathbf{m}_{2}, ..., \mathbf{m}_{n^{l}}\}$, which is called a ``layer-wise pool'' or simply a ``pool'' in our formulation. Let's consider that we need to learn $d^l$ groups of weights for this layer, denoted by $\mathbf{W}^{l} =\{\mathbf{w}_{1}, \mathbf{w}_{2}, ..., \mathbf{w}_{d^l}\}$. 
NWS builds this layer by searching for an optimal combination ($\boldsymbol{comb}$) of weights $\mathbf{K}^{l} =\{\mathbf{k}_{1}, \mathbf{k}_{2}, ..., \mathbf{k}_{d^l}\}$ where $\mathbf{k}_{i} \in \mathbf{M}^{l}$. Note that the search processes of different layers are conducted simultaneously with the objective $
\arg \min_{\boldsymbol{comb}} \mathcal{L}(f_\mathbf{K}(\mathbf{x}), \mathbf{y})
$. Because new models are formed with indexed grouped weights, if there are a number of models, NWS can greatly reduce memory cost by saving their combination indices as well as the search space. In this paper, we group weights in the form of convolution kernels, e.g., for a convolution kernel with size $3\times 3$, $9$ weight values are grouped together. We have demonstrated the effectiveness of our approach with varying convolutional network architectures. 

To the best of our knowledge, neural weight search as introduced in this paper has not been explored before. NWS is completely different from both `kernel search' in mixed-integer linear programming and `kernel search optimization' used in kernel methods like SVM. NWS is analogous to Neural Architecture Search (NAS). NAS designs a search space of model components beforehand and obtains an optimal model architecture by evaluating the combinations of these components. NWS designs a search space for neural weights and \textit{automatically searches for an optimal combination of weights to build a network}. 


\section{Proposed Method}
\label{sec:method}
Our proposed method has two main components, namely, Neural Weight Search (NWS) algorithm (Section~\ref{subsec:kernel_searching}), and design of the search space (Section~\ref{subsec:kernel_market_initilization}). 
\subsection{NWS Algorithm}
\label{subsec:kernel_searching}
As mentioned in Section~\ref{subsec:nws}, we divide weights into groups such that each group represents a convolution kernel. We define a convolution kernel as a $k \times k$ matrix in the float domain. To build a convolution layer $l$, we search a layer-wise pool for an optimal combination of kernels, $\mathbf{K}^{l} \in \mathbb{R}^{d^{l} \times k \times k}$, where $d^{l}=d^{l}_{in} * d^{l}_{out}$ and $d^{l}_{in}$, $d^{l}_{out}$ are the number of input channels and output channels respectively. A layer-wise pool $\mathbf{M}^{l} \in \mathbb{R}^{n^{l} \times k \times k}$ is learned and fixed during the pretraining stage, where $n^{l}$ is the number of kernels in the pool (see Section~\ref{subsec:kernel_market_initilization}). The kernels in the pool are indexed by a non-negative integer number ranging from $0$ to $n^{l}-1$, so the pool $\mathbf{M}^{l}$ is a \textbf{lookup table} that can return corresponding kernels by giving indices.  
Since the kernels in the pool are indexed, $\mathbf{K}^{l}$ can be saved as a vector $\mathbf{C}^{l}$ (whose values are non-negative integers), $\mathbf{C}^{l} \in \mathbb{N}^{d^{l}}$, and $\mathbf{C}^{l}$ can be easily mapped to the float domain $\mathbf{K}^{l}$ by doing lookup operations in $\mathbf{M}^{l}$. Please see Figure~\ref{fig:KP}. 

One straightforward way to search the layer-wise pools is using brute force search based on classification performance. However, per layer, the number of possible combinations of kernels to be searched for is $(n^{l})^{d^{l}}$ and $d^{l}$ is a constant given a model. 
Consequently, it is not feasible to try every possible combination using brute force search as the search space is extremely large. Hence, we introduce an efficient, end-to-end searching algorithm for finding the optimal combination of kernels for making up layers and a model as follows. We utilise an auxiliary component that is a sequence of learnable temporary kernel weights $\mathbf{W}^{l} \in \mathbb{R}^{d^{l} \times k \times k}$ with the same size of $\mathbf{K}^{l}$. A layer needs to be filled with kernels selected from the pool and $\mathbf{W}^{l}$ can be regarded as placeholders. Each temporary kernel is replaced by a kernel in the pool based on a similarity metric such as $L2$ distance. Kernel weights selected from the pool form the model and are used to process inputs, resulting in classification loss. Temporary kernels are then updated by the classification loss and the similarity metric loss. We repeat this process over multiple iterations to update the temporary kernels such that an optimal set of $\mathbf{K}^{l}$ can be found from the pool given a new task. Temporary kernel weights can be initialised using the weights mapped from kernel indices of a previous task.

To wrap up, we search the pools for the optimal combination of kernels for each layer in the model simultaneously. The formed layers then process the input in forward propagation. However, in backpropagation, the gradients of $\mathbf{K}^{l}$ induced by the classification loss are used to update the temporary kernel weights $\mathbf{W}^{l}$, and the kernel weights in the pool remain fixed. As given in Eq.~\ref{eq:NWS}, searching the weights of a kernel is the nearest neighbour search problem, where $\mathbf{k}$ and $\mathbf{w}$ are the selected kernel and the temporary kernel, respectively: 
\begin{equation}
\label{eq:NWS}
\mathbf{k}_{i}=\text{NWS}(\mathbf{w};\mathbf{M}), \text { where } i=\arg \min _{i}\left\|\mathbf{w}-\mathbf{k}_{i}\right\|_{2}.
\end{equation}
Since NWS is a non-differential operation, passing the gradients of selected kernels to temporary kernels is not straightforward and we use Straight-Through Estimation (STE) for this purpose. The loss can be then defined as 
\begin{align}
\mathcal{L}_{\text{NWS}}(x, y, \mathbf{W}, \mathbf{M})) = & -\sum_{j}^{V} \delta_{y=j}\log(p_{j}(x))   \nonumber  \\
&+\|s g[\text{NWS}(\mathbf{W;M})]-\mathbf{W}\|_{2}^{2} \label{eq:NWS_loss}, 
\end{align}
where the first term in Eq.~\ref{eq:NWS_loss} is a softmax cross-entropy loss, $\delta_{y=j}$ is an indicator function, and $p_{j}(x)$ denotes the prediction logits of the $j_{th}$ class (V classes in total). The second term in Eq.~\ref{eq:NWS_loss} is a similarity loss (mean squared error is used in practice). Stop-gradient, $sg[.]$, is an operation that prevents gradients to propagate to its argument. 
As shown in Figure~\ref{fig:KP}, gradients of the two aforementioned losses propagate back to the temporary kernels: (1) the direct gradients of the similarity loss; and (2) the indirect STE gradients of the classification loss.

There are a couple of points to note. For each layer, we use a separate pool. Our investigation has shown that a single shared pool for all layers performs worse than layer-wise pools. Because during the training or inference temporary kernels are not directly used for the forward propagation, they are never stored and can be discarded after the model is built. The whole training and inference process is presented in Algorithm~\ref{alg:KP}. 


\begin{algorithm}[!ht]
\caption{Task Incremental Learning with Neural Weight Search}
\label{alg:KP}
 \SetKwInOut{Input}{Input}
 \SetKwInput{Require}{Require}
  \SetKwInOut{Initial}{Initialise}

 \Require{ 
 \;
 $\mathbf{C}_{0}$  $\triangleright$ kernel indices of the pretrained model \;
 $\mathbf{M}$ $\triangleright$ pretrained pools \;
 NWS $\triangleright$ neural weight search function that takes temporary kernels as inputs \;
  $\mathcal{E}$ $\triangleright$ embedding function that encodes input kernels into corresponding non-negative integers via a pool \;
  $\mathcal{D}$ $\triangleright$ lookup operation that returns corresponding kernels given indices via a pool (lookup table) \;
 }
\For{task $ t=1 ; t<= N$ }{
    get task-specific data   $\mathbf{x}_{t}$, $\mathbf{y}_{t}$ \;
  $\mathbf{W}_{t}$  $\leftarrow$ $\mathcal{D}$($\mathbf{C}_{t -1}$; $\mathbf{M}$ ) 
 
     \For{layer $l=1 ;l<= L$ }{
        $\mathbf{K}_{t}^{l}$, $\text{diff}_{t}^{l}$  $\leftarrow$ {NWS}($\mathbf{W}_{t}^{l};{M}^{l}$ ) \;
        $\mathcal{L}_{\text{diff}}$ $\leftarrow$  $\mathcal{L}_{\text{diff}}$ + $\text{diff}_{t}^{l}$ \;
       \uIf{$l=1$}{
              $\mathbf{o}_{t}^{l}$  $\leftarrow$  f($\mathbf{x}_{t}; \mathbf{K}_{t}^{l}$ ) \;
            }
      \Else{
        $\mathbf{o}_{t}^{l}$  $\leftarrow$  f($\mathbf{o}_{t}^{l-1}; \mathbf{K}_{t}^{l})$ 
      } 
     }
  $\mathcal{L}_{ce}$ $\leftarrow$ $min_{\mathbf{W}_{t}} \mathcal{L}(o_{t}^{L}, y^{t}) $ \;

 $\mathbf{W}_{t}$  $\leftarrow$  UPDATE($\mathbf{W}_{t}$; $\mathcal{L}_{ce}, \mathcal{L}_{\text{diff}}$) $\triangleright$ update temporary kernels based on Eq.~\ref{eq:NWS_loss} \;
  
   $\mathbf{C}_{t}$  $\leftarrow$  $\mathcal{E}$ ($\mathbf{W}_{t}$; $\mathbf{M}$ ) $\triangleright$ save model as indices \;
   
   $\mathbf{K}_{t}$  $\leftarrow$  $\mathcal{D}$ ($\mathbf{C}_{t}$; $\mathbf{M}$) $\triangleright$ During inference, map indices back to kernel weights by indexing in pools\;
    Excute INFERENCE$(\mathbf{x}_{t}; \mathbf{K}_{t})$ 
}
\vspace{-0.2cm}
\end{algorithm}

\subsection{Design of Search Space}
\label{subsec:kernel_market_initilization}
There could be many ways to design layer-wise search spaces (pools). One straightforward way is pretraining a model and using all kernel weights as the search space. However, in this case, as the size of this search space is proportional to the number of input and output channels of a layer, large numbers can bring about significant search costs. To tackle this, we propose a novel knowledge distillation strategy that can distil from a network a compact search space {\bf with a predefined pool size}. Constructing the pools can be achieved by minimising the following loss function: 
\begin{equation}
\label{eq:pool_loss}
\min_{\mathbf{W},\mathbf{M}}\mathcal{L}_{\text{KP}}(\mathbf{x}, \mathbf{y}, \mathbf{W}, \mathbf{M}) =  \min_{\mathbf{W}}\mathcal{L}_{\text{NWS}}+\beta\min_{\mathbf{M} } \mathcal{L}_{\text{WD}},
\end{equation}
where the $\mathcal{L}_{NWS}$ is the weight search loss for updating temporary kernels used in Section~\ref{subsec:kernel_searching}. We define the \textit{weight distillation} (WD) loss as follows:
\begin{equation}
\mathcal{L}_{\text{WD}} = \|s g[\mathbf{W}]-\text{NWS}(\mathbf{W;M})\|_{2}^{2}. 
\label{eq:distillation_loss}
\end{equation}
In Eq.~\ref{eq:pool_loss}, the first term optimises the temporary weights to find the optimal selection of kernel weights in the pool. 
The second term, i.e., weight distillation loss allows the selected kernel weights (that are trainable in this stage) from the pools to be updated, bringing them closer to corresponding temporary weights. A coefficient $\beta$ is used to control the speed for updating the pools. The construction of the layer-wise pools (weight embedding space) is analogous to that of the feature embedding space discussed in VQ-VAE~\cite{VQ_VAE}, which is trivial to implement. The pretraining and distillation are conducted simultaneously in a large-scale dataset like ImageNet~\cite{DS_ImageNet} to ensure the generalization.

Once pretraining is completed, for each layer, the kernel weights in pools are frozen and temporary weights are discarded. The resulting pools can then be utilized for building novel models and learning new tasks as discussed in Section~\ref{subsec:kernel_searching}. 

\subsection{Implementation Details}
Our NWS algorithm is a model-agnostic learning method, and it is implemented by simply replacing the convolution layers (including short-cut layers) with NWS-incorporated convolution layers. An NWS-incorporated convolution layer has temporary kernels (which are discarded after searching) and a pool. Our investigation has shown that the optimal number of kernels for each pool is $n^l=512$ (see App.~\ref{app:ablation} for details). To evaluate our approach, we present results with different architectures including ResNet-18, ResNet-34~\cite{Resnet}, MobileNetV2~\cite{mobilenetv2}, and VGG~\cite{VGG}. For instance, for ResNet-18~\cite{Resnet}, we replace the last fully connected layer with a convolution layer (kernel size of $1\times 1$). The kernel size ($k \times k$) for the remaining layers is set based on the default parameters of ResNet-18. 

The overall loss function in Eq.~\ref{eq:pool_loss} is used to construct layer-wise pools by pretraining the model such as ResNet-18 on a large-scale image dataset, i.e., ImageNet~\cite{DS_ImageNet}, where we randomly initialise the kernels in the pools as well as temporary kernels. Then, during the incremental learning, layers of a new model are built simultaneously with the NWS loss (Eq.~\ref{eq:NWS_loss}) in an end-to-end manner.

\begin{figure*}[t]
\centering
\includegraphics[width=0.8\textwidth]{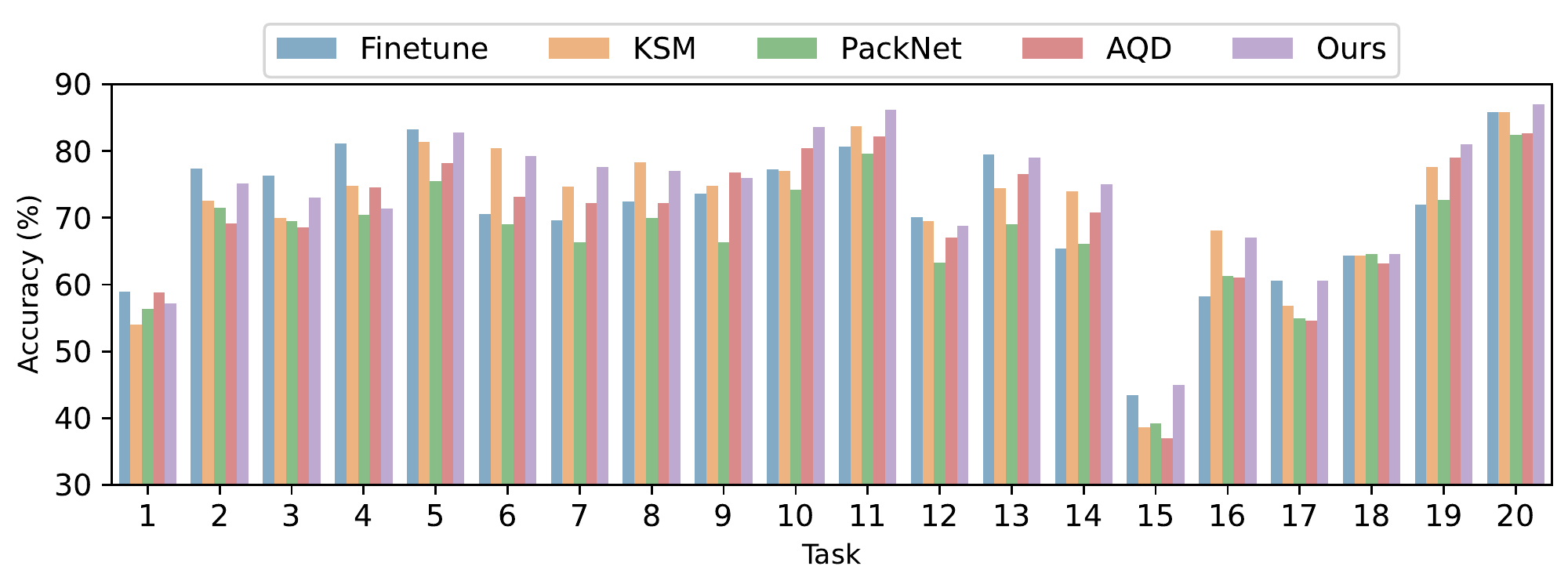} 
\caption{Comparison of methods described in Section~\ref{subsec:baselines} in terms of task-wise accuracy on Split-CIFAR-100.}
\label{fig:task_acc_cifar}
\end{figure*}

\begin{figure*}[t]
\begin{center}
\includegraphics[width=0.9\textwidth]{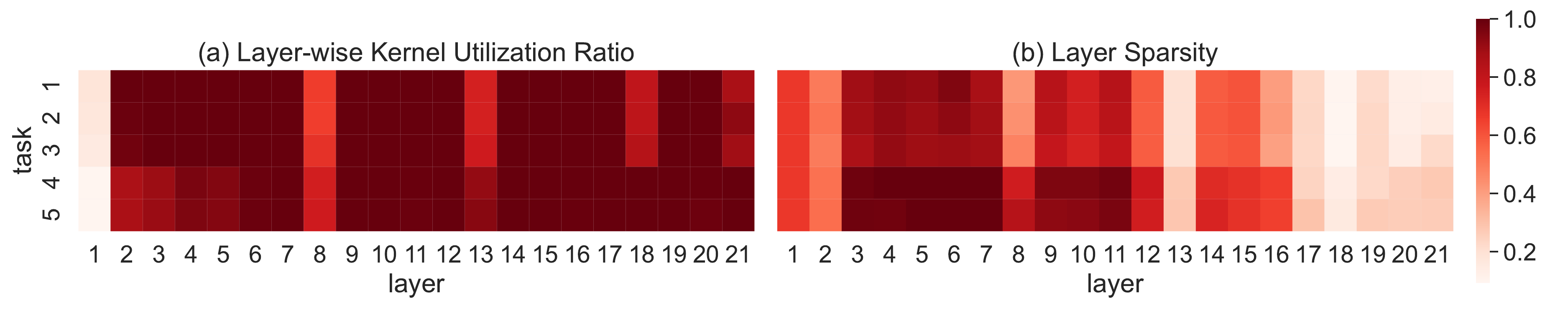} 
\caption{Heat map visualisation of (a) layer-wise kernel utilization ratio, and (b) layer sparsity obtained on the CUB-to-Sketches benchmark (denoted as task $1$ to task $5$ respectively).}
\label{fig:heatmap}
\end{center}
\vspace{-0.2cm}
\end{figure*}

\section{Experiments}
\subsection{Benchmarks}
\label{sub:benchmarks}
Following the previous works~\cite{CL_KSM_yang2021,CL_PackNet}, we evaluate our proposed methods on two existing TIL benchmarks: Split-CIFAR-100, and CUB-to-Sketches. The Split-CIFAR-100 benchmark contains $20$ tasks, where each is a $5$-way classification task with an image size of $32 \times 32$. The split is done in a way that the $5$ classes in a task have the same superclass~\cite{Cifar100}. The CUB-to-Sketches benchmark contains $5$ datasets, each treated as a task, including CUB-200~\cite{DS_CUB200}, Cars-196~\cite{DS_STANFORD_CARS196}, Flowers-102~\cite{DS_OXFORD_FLOWERS102}, WikiArt-195~\cite{DS_WIKIART195}, and Sketches-250\cite{DS_SKETCH250}. Here, for instance, CUB-200 means the CUB dataset has $200$ classes, and so on, and for these datasets, the images are resized to $224 \times 224$.

\subsection{Evaluation Metric} 
We report task-wise accuracy by evaluating each task individually with the corresponding task-wise model. We also use \textit{average accuracy} over all tasks, which is defined as:
$
\overline{\mathcal{A}}=\frac{1}{T} \sum_{i=1}^{T} A_{i},
$
where $T$ is the total number of tasks. Following ~\cite{CL_CPG_hung2019,CL_KSM_yang2021, CL_PackNet} total memory cost is reported which includes the backbone model and introduced masks as well as statistics (means and variances) of batch normalizations per task for baselines or includes the shared layer-wise kernel pools and individual kernel indices as well as statistics of batch normalizations per task for our method.

\subsection{Baselines}
\label{subsec:baselines}
We compare our algorithm with the following baseline methods: (1) \textbf{Finetune}. It finetunes a pretrained model for each task separately. It theoretically provides the upper bound of accuracy and uncontrollable memory growth. (2) \textbf{KSM}~\cite{CL_KSM_yang2021}. It fixes the pretrained backbone model and learns soft kernel-wise masks for each task. (3) \textbf{PackNet}~\cite{CL_PackNet}. It learns binary kernel-wise masks for each task and updates the pretrained model. After each task is learned, it frees up a fixed ratio of model weights and only released weights are learnable for the next task. (4) \textbf{AQD}~\cite{MC_AQD_liu2020}. It quantises weights as well as features when finetuning the pretrained model in an element-wise manner for each task individually. 

\subsection{Experimental Setup} 
\label{subsec: Experimental Setup}
For a fair comparison, we use the same backbone model architecture (i.e., ResNet-18~\cite{Resnet}) and the same common hyperparameters for baselines and our method. For method-specific hyperparameters, default values are chosen in their original implementation (See App.~\ref{app:baseline} for details). We run each experiment with $3$ different seeds and report the average results.
\\
\textbf{Pretraining.} For all baselines, we used an initial ResNet-18 pretrained on ImageNet~\cite{DS_ImageNet} (from Pytorch model zoo). For NWS, we pretrain our pools on ImageNet for $160$ epochs. Each layer-wise pool has $512$ kernels. Both kernels in the pools and temporary kernels are randomly initialised during the pretraining. To learn the task $t$, our method initialises the temporary weights from the previous task $t-1$ by looking up the indices for real-valued kernels. 
\\
\textbf{Hyperparameters.} Following~\cite{CL_KSM_yang2021,CL_CPG_hung2019}, all models are trained with a stochastic gradient descent (SGD) optimizer with $0.9$ momentum and $1e^{-5}$ weight decay in $100$ epochs. In the case of Split-CIFAR-100, the initial learning rate is set to $0.01$. In the case of CUB-to-Sketches, the initial learning rate is set to $0.001$. The learning rate is divided by $10$ after $50$ epochs and $80$ epochs. For NWS, we empirically set $\beta$ to $0.5$ ($0.1$) in Eq.~\ref{eq:pool_loss} for Split-CIFAR-100 (CUB-to-Sketches). We report further results conducted with different hyperparameters in App.~\ref{app:hyperparameters}. 

\begin{table}[!ht]
 \rowcolors{2}{gray!10}{gray!40}
  \centering 
   \caption{Comparison of methods described in Section~\ref{subsec:baselines} in terms of average classification accuracy and total memory cost on Split-CIFAR-100.} 
  
     \begin{tabularx}{0.45\textwidth}{@{\extracolsep{\fill}} l c c  @{}} 
    \toprule
    \rowcolor{white}
    Method  & Avg Acc (\%) &  Memory (MB) \\ [0.5ex] 
    \hline 
    Finetune & 71.3 &  892.0
    \\
    KSM~\cite{CL_KSM_yang2021} &  71.5  & 192.5 
    \\
    PackNet~\cite{CL_PackNet} & 67.1 & 55.2
    \\
    AQD~\cite{MC_AQD_liu2020} & 69.9  & 52.7
  
    \\
    Ours & \textbf{73.4} & \textbf{33.9}  
  
    \\
     \bottomrule
    \end{tabularx}
    \label{tab:avg_acc_cifar} 
\end{table}

\begin{table*}[!ht]
 \rowcolors{6}{gray!10}{gray!40}
    \small \centering
    \caption{Comparison of methods described in Section~\ref{subsec:baselines} in terms of task-wise accuracy on CUB-to-Sketches benchmark.}

    \begin{tabular*}{0.9\textwidth}{@{\extracolsep{\fill}} l cccccc c @{}}
    \toprule
    \multirow{3}{*}{Method} & \multicolumn{6}{c}{Task-wise Accuracy (\%)} & \multirow{3}{*}{Memory (MB)} \\ \cmidrule(lr){2-7}  & Task1 & Task2 & Task3 & Task4 & Task5 &  \multirow{2}{*}{Avg}  \\
   \cmidrule(lr){2-6}           & CUB           & Cars          & Flowers       & WikiArt & Sketch   \\ \midrule
    Finetune                    & 77.4          & 84.1          &    94.5       & 74.2    & 76.9  & 81.4  & 223.0
 \\ KSM~\cite{CL_KSM_yang2021}  & 65.9          & 79.7          & \textbf{93.5} & 66.2    & 73.9  & 75.8  &  76.9
    \\PackNet~\cite{CL_PackNet} & \textbf{78.4} & 82.4          & 90.7          & 69.0    & \textbf{75.3}  & 79.1  &  56.0
 \\AQD~\cite{MC_AQD_liu2020}    & 43.0	        & 51.8          & 63.4          & 60.5    &	69.7  & 57.7  & 13.7

  \\Ours                        & 77.0 & \textbf{87.8} & 93.0  & \textbf{73.9} &75.1 &  \textbf{81.3}  & \textbf{9.9} \\ 
    \bottomrule
    \end{tabular*}
    \label{tab:CUB-to-Sketches}
\end{table*}

\subsection{Experimental Results}
In terms of accuracy, Figure~\ref{fig:task_acc_cifar} compares task-wise inference accuracy on Split-CIFAR-100, where our method outperforms all the methods in 8 tasks out of 20, including the Finetune. PackNet performs similarly to KSM from task $2$ to $4$; but for the latter tasks, its performance is worse than other baselines, supporting the phenomenon that learnable weights run out after a certain point. AQD performs slightly worse than KSM and Finetune; because network quantisation techniques limit the learning of rich representations. Looking at Table~\ref{tab:avg_acc_cifar}, our method outperforms KSM by a margin of $1.9\%$ in terms of average accuracy. Further results on CUB-to-Sketches are presented in Table~\ref{tab:CUB-to-Sketches}. PackNet achieves slightly better results than KSM, except for WikiArt. This might be due to the that the number of tasks in CUB-to-Sketches is much smaller than Split-CIFAR-100 ($5$ vs. $20$). AQD performs much worse than others. We conjecture that it is because CUB-to-Sketches has higher resolution images as compared to Split-CIFAR-100 ($224^{2}$ vs. $32^{2}$). 
Excluding Finetune which is memory-unfriendly, overall, our method achieves the best average accuracy. It performs better than other methods for the tasks, Cars and WikiArt, and is on par with the best performing baselines for the CUB ($-1.4\%$), Flowers ($-0.5\%$) and Sketch ($-0.2\%$).

In terms of memory, a similar trend can be observed for both Split-CIFAR-100 and CUB-to-Sketches. Looking at the Tables~\ref{tab:avg_acc_cifar} and~\ref{tab:CUB-to-Sketches}, as compared to Finetune, PackNet, KSM and AQD, our method respectively saves up to $96\%$, $39\%$ $82\%$, $35\%$ on Split-CIFAR-100 benchmark and $95\%$, $82\%$, $87\%$, $28\%$ memory on CUB-to-Sketches benchmark. 

In terms of running time during the inference, all models excluding AQD are approximately the same. NWS can map saved kernel indices of a model to the weights of the model with negligible time thanks to the lookup operation. Taken together, our proposed method achieves competitive performance on two TIL benchmarks and demonstrates an unattainable memory saving as compared to the baselines~\footnote{We also compared with CPG~\cite{CL_CPG_hung2019}. However, our investigation showed that if ResNet-18 is used as the backbone on Split-CIFAR100, CPG becomes sensitive to predefined thresholds that are used to control the training and tends to fail or grow too much, which led to poor performance for this method and therefore it was excluded for fairness.}
\subsection{Different Model Architectures}
\label{subsec:arch}
To demonstrate our approach can be generalised, we test NWS algorithm with another 3 different architectures: Resnet-34~\cite{Resnet}, MobileNet-V2~\cite{mobilenetv2}, VGG-16~\cite{VGG}. We compare NWS-incorporated models with baselines in Table~\ref{tab:res32_cifar}. The baselines finetune a corresponding pretrained model (pretrained on ImageNet) for each task separately. Please refer to App.~\ref{app:arch} for pretraining and training setups. Results show that our NWS is compatible with modern deep neural networks and NWS-incorporated networks can offer a large memory reduction with competitive inference accuracy. Note that NWS-VGG16 achieves a higher compression rate ($97.5\%$) than that of NWS-MobileNetV2 ($81\%$). It is because $1\times1$ kernels are extensively used in MobileNetV2 and from the model compression perspective for a $1\times1$ kernel NWS only compresses $1$ float value into $1$ integer value.

\begin{table}[!htb]
 \rowcolors{6}{gray!10}{gray!40}
  \centering 
  \caption{Comparison of different architectures on Split-CIFAR-100. } 
     \begin{tabularx}{0.45\textwidth}{@{\extracolsep{\fill}} l c c   c@{}}  
    \toprule
     \rowcolor{white}
    Method  & Avg Acc (\%) & Memory (MB) \\ [0.5ex] 
    \hline 
     Finetune-Res34     &  77.0 & 1,628.0
  \\ Finetune-VGG16     &  75.6  & 1,124.0
  \\ Finetune-MobilnetV2 &  75.7 & 272.0

  \\ NWS-Res34      & 74.8 & 59.6
  \\ NWS-VGG16      & 74.8 & 28.0
  \\ NWS-MobileNetV2 & 71.5 & 52.6

\\
     \bottomrule
    \end{tabularx}
    \label{tab:res32_cifar} 
\end{table}

\subsection{Analysis of Selected Kernels}
\label{subsec:analysis}
In this section, we provide further insight into how kernels are used for different layers. For this purpose, we introduce two new concepts, namely, \textit{layer-wise kernel utilization ratio} and \textit{layer sparsity}.

We define the \textit{layer-wise kernel utilization ratio} (KUR) as $\text{KUR} = { U^{l}  / n^{l}}$, where $U^{l}$ is the number of unique selected kernels for the layer $l$ and $n^{l}$ is the number of kernels in the pool ($512$ kernels in our case). We compute KUR on the CUB-to-Sketches benchmark. The ResNet-18 has a total of 21 layers (including short-cut layers). Figure~\ref{fig:heatmap}-(a) shows the layer $1$ and $8$ utilise a smaller number of unique kernels from the corresponding layer-wise pool. For layer $1$ (the first layer), very low utilization ratios are observed. A larger utilization ratio indicates most of the kernels in the pool are selected. Kernels in the first layers capture coarse common local features (e.g., line, curve, and dot), whereas kernels in the following layers extract fine-grained specialized global features (e.g., ear, eye, and head). Therefore, we conjecture that diversity (large KUR) is necessary for kernels in the subsequent layers to ensure specialisation as compared to the first layers. 

We also investigate the \textit{layer sparsity}. Intuitively, the more a kernel is reused in a layer, the more important it is. A kernel selected a few times is less important in contrast; therefore, setting its weight values to zero may hardly hamper the performance, which can be used as a means for network sparsification. We denote the selection times of a unique index as $h_{u}^{l}$ and an adaptive layer-wise threshold as $\sqrt{d^{l}}$ where ${d^{l}}$ is the number of required kernels to build the layer. We formulate the layer sparsity as follows: 
\begin{equation}
\text{LS}^{l} =  \sum^{U^{l}}_{u=1}{h_{u}^{l} \over d^{l}}  \quad
\textrm{s.t.} \quad  h_{u}^{l} < \sqrt{d^{l}}.
\label{eq:LS}
\end{equation} 
Another heatmap is used to visualise the layer-wise sparsity for the CUB-to-Sketches benchmark. As shown in Figure~\ref{fig:heatmap} (b), we observe higher levels of layer sparsity in the latter tasks (i.e., tasks 4 and 5). It might be that the tasks are learned in a sequential manner (temporary kernels of the current task are initialised with the reconstructed kernel indices of the previous model) and sparsity might have been inherited. 

\section{Conclusion}
In this paper, we propose a novel method called Neural Weight Search for task incremental learning. Our algorithm learns new models by searching grouped weights saved in layer-wise pools and saves learned models in the form of indices, which significantly reduces the memory cost. NWS is an out-of-the-box mechanism that can be easily integrated with modern deep learning methods. Our experiments show NWS achieves state-of-the-art performance on the Split-CIFAR-100 and CUB-to-Sketches benchmarks in terms of both accuracy and memory.


\footnotesize{
\section*{Acknowledgements}
\noindent 
The work of Jian Jiang has been supported by the King's China Scholarship Council (K-CSC) PhD Scholarship programme and NVIDIA Academic Hardware Grant Program. The work of Oya Celiktutan has been supported by the LISI Project funded by the EPSRC UK (Grant Ref: EP/V010875/1). Finally, this work has been partially supported by Toyota Motor Europe (TME) and Toyota Motor Corporation (TMC). 
}
\clearpage

\begin{appendices}

In this appendix we present supplementary materials as follows:
Insights into Neural Weight Search are presented in Section~\ref{app:NWS}. Ablation studies are shown in Section~\ref{app:ablation}. Experiments of our method conducted with different hyperparameters are shown in Section~\ref{app:hyperparameters}. Training setup for different network architecture are in Section~\ref{app:arch}. Implementation details of the baselines used in the main paper are discussed in Section~\ref{app:baseline}. Visualisations of distributions of kernel indices are shown in Section~\ref{app:visual_indices}. 

\section{Insights into Neural Weight Search}
\label{app:NWS}
In transfer learning~\cite{Transfer_ref1,Transfer_ref2}, weights of layers from a pretrained model can be reused for finetuning new models. Considering the training cost, one efficient way is freezing parts of weights and finetuning the rest, e.g., fixing the low-level weights and retraining the high-level weights and vice versa. However, conducting finetuning on all weights can usually increase learning ability. Intuitively, it is inevitable that some weights of good generalization ability are updated during the finetuning and updated weights may be not suitable for finetuning other models. 

Given a set of well-generalized weights, is it possible to utilize them to build a model without changing their values? Motivated by this, we propose a new problem setting named Neural Weight Search (NWS). Distinctive from conventional finetuning strategies, {\bf NWS aims to search optimal combinations of frozen weights from a search space with repetition to build different models instead of changing the values of weights}. In this way, the re-utilization of weights is maximized, resulting in a large memory reduction. Nonetheless, we need to take care of the computation cost of the search process which is influenced by two main factors: (1) the size of the search space. (2) the searching strategy. Note that, the computation workload and the running time of a searched model during the inference are the same as those of a vanilla finetuned model. We elaborate on how to design the search space in the rest of this section.

{\bf \textit{Search Space:}} We recommend using grouped weights as elements to reduce the size of the search space because treating each weight scalar as an element will induce a search space of large size. In the main paper, we focus on convolution networks so we can group weights in convolution kernels, e.g., for a kernel with size $3\times 3$, $9$ weight values are grouped together. In terms of a fully connected (FC) layer, although its weights cannot be grouped into kernels, it is also feasible to divide weights into groups with different strategies. For instance, given an FC layer with input size $4$ and output size $6$, weights can be divided into $6$ groups with each group having $4\times1$ weight scalars or $4$ groups with each group having $2 \times 3$ weight scalars. We leave this direction to readers.

\section{Ablation Studies}
\label{app:ablation}
We study the individual impact of two modifications, i.e., pretrained pools, and initialisation of temporary kernels, on the performance of our proposed method using the Split-CIFAR-100 dataset. More explicitly, we conduct the following experiments: 1) We evaluate the importance of the generalization of weights in pools by pretraining the NWS using a subset ($100$ classes) of ImageNet (NWS-subImg) and the original version (Original) is pretrained using the full set. 2) We use randomly initialized temporary weights for each new task (NWS-random). As shown in Table~\ref{tab:abl_cifar}, pretraining NWS with more classes (Original vs. NWS-subImg) on ImageNet benefits incremental learning, implying that more data helps NWS learn more distinctive and generalized weights. NWS-random has the worst average accuracy, demonstrating the importance of the initialisation of temporary kernels.

\begin{table*}[!ht]
    \rowcolors{6}{gray!10}{gray!40}
  \centering 
  \caption{Ablation studies on the Split-CIFAR-100 dataset. NWS-subImg: NWS pretrained using a portion of the ImageNet. NWS-random: NWS with randomly initialised temporary kernels.} 
     \begin{tabularx}{0.9\textwidth}{@{\extracolsep{\fill}} l c c c c @{}} 
    \toprule
    
      \rowcolor{white}
    Method  & Average Accuracy (\%) & Per Task (MB) & Assist (MB) & Total (MB)  \\ [0.5ex] 
    \hline  Original  & \textbf{73.4} &       1.6 & 1.3 & 33.9   
    \\ NWS-subImg  & 72.5 &   1.6    & 1.3  &  34.1
    \\ NWS-random  & 68.7 &   \textbf{1.5}    & 1.3 & \textbf{32.0}
    \\
     \bottomrule
    \end{tabularx}
    \label{tab:abl_cifar} 
\end{table*}

\section{Different Hyperparameters}
\label{app:hyperparameters}
We evaluate our method from several aspects: \textit{optimizer}, \textit{learning rate (LR)}, \textit{beta} (the coefficient in the Eq. 3 in the main paper), and~\textit{random seed}. For stochastic gradient descent (SGD), we used the same setting as that used in our main paper. More specifically, we use $0.9$ as momentum, 4e-5 as weight decay, and set `Nesterov' to `True' for SGD. For Adam optimizer, we use Pytorch default setting to initialise. Milestones are used to divide the learning rate by $10$ at certain epochs. We use milestones $\{50, 80\}$ when SGD optimizer is used and $\{50\}$ for Adam optimiser. For example, an initial learning rate $0.1$ with milestones $\{50, 80\}$ will become $0.01$ after epoch $50$ and then become $ 0.001$ after epoch $80$. Results are reported in Table~\ref{tab:hyper_cifar} and Table~\ref{tab:hyper_CUB}. We also calculate the mean and standard deviation of average accuracy and total memory. Low standard deviations show our model is stable.

We further investigate the impact of the size of a kernel pool. We pretrain kernel pools with different sizes ($128$, $256$, $512$ and $1024$) on Sub-ImageNet. We obtain average results on $3$ different seeds in Tab.~\ref{tab:size_kp}. Results show that the size of $512$ achieves the best performance.
\begin{table*}[!htb]
 \rowcolors{6}{gray!10}{gray!40}
  \centering 
  \caption{Hyperparameters on split-CIFAR-100. `LR' refers to the learning rate; `Beta' refers to the coefficient; `Seed' refers to the random seed; `Memory' refers to the total memory cost including saving the kernel pools; `Mean' (`STD') refers to the mean (standard deviation) of average accuracy or total memory.} 
     \begin{tabularx}{0.9\textwidth}{@{\extracolsep{\fill}} l c c c c c c @{}} 
    \toprule
      \rowcolor{white}
    Optimizer  & LR & Beta& Epoch &  Seed& Average Accuracy (\%) & Memory  \\ [0.5ex] 
    \hline 
       SGD  & 1e-1 & 0.5 & 100   & 1993 & 73.4 & 32.4
    \\ SGD  & 1e-1 & 0.5 & 100   & 1994 & 73.9 & 32.7
    \\ SGD  & 1e-1 & 0.5 & 100   & 1995 & 73.8 & 32.6
     \\SGD  & 1e-2 & 0.1 & 100   & 1993 & 72.9 & 33.8
    \\ SGD  & 1e-2 & 0.1 & 100   & 1994 & 72.5 & 33.9
    \\ SGD  & 1e-2 & 0.1 & 100   & 1995 & 74.7 & 33.9
     \\SGD  & 1e-2 & 0.5 & 100   & 1993 & 74.1 & 33.9
    \\ SGD  & 1e-2 & 0.5 & 100   & 1994 & 72.9 & 33.9
    \\ SGD  & 1e-2 & 0.5 & 100   & 1995 & 73.4 & 33.9
     \\SGD  & 1e-2 & 1   & 100   & 1993 & 75.6 & 33.8
    \\ SGD  & 1e-2 & 1   & 100   & 1994 & 72.6 & 33.9
    \\ SGD  & 1e-2 & 1   & 100   & 1995 & 70.2 & 34.0
     \\SGD  & 1e-3 & 0.5 & 100   & 1993 & 79.8 & 33.6
    \\ SGD  & 1e-3 & 0.5 & 100   & 1994 & 79.7 & 33.6
    \\ SGD  & 1e-3 & 0.5 & 100   & 1995 & 79.3 & 33.6
    \\ \hline 
      \rowcolor{white}
       Mean &      &     &       &      & 74.6  & 33.6
         
    \\   \rowcolor{white}
    STD  &      &     &       &      & 2.85   & 0.54
    \\ \hline \hline
      Adam   & 1e-2 & 0.5 &  100  & 1993  &  71.0 & 25.4
    \\Adam   & 1e-2 & 0.5 &  100  & 1994  &  71.3 & 25.6
    \\Adam   & 1e-2 & 0.5 &  100  & 1995  &  71.9 & 25.7

    \\ \hline 
      \rowcolor{white}
      Mean &      &     &       &      & 71.6  & 25.6
    \\   \rowcolor{white}
    STD &       &     &       &      & 0.46  & 0.15

  \\
     \bottomrule
    \end{tabularx}
    \label{tab:hyper_cifar} 
\end{table*}

\begin{table*}[!htb]
 \rowcolors{6}{gray!10}{gray!40}
  \centering 
  \caption{Hyperparameters on CUB-to-Food. `LR' refers to the learning rate; `Beta' refers to the coefficient; `Seed' refers to the random seed; `Memory' refers to the total memory cost including saving the kernel pools; `Mean' (`STD') refers to the mean (standard deviation) of average accuracy or total memory.} 
     \begin{tabularx}{0.9\textwidth}{@{\extracolsep{\fill}} l c c c c c  c@{}} 
    \toprule
      \rowcolor{white}
    Optimizer  & LR & Beta & Epoch & Seed& Average Accuracy (\%) & Memory  \\ [0.5ex] 
    \hline 
     SGD  & 1e-3 & 0.1 & 100  &  1993 &  81.7 & 9.9
  \\ SGD  & 1e-3 & 0.1 & 100  &  1994 &  81.0 & 9.9
  \\ SGD  & 1e-3 & 0.1 & 100  &  1995 &  81.4 & 9.9

\\ \hline 
  \rowcolor{white}
       Mean &      &     &       &      & 81.4  & 9.9
    \\   \rowcolor{white}
    STD  &      &     &       &      & 0.44   & 0.0
    \\ \hline \hline
    Adam   & 1e-3 & 0.1 &  100   & 1993 & 74.9 & 7.6
  \\ Adam  & 1e-3 & 0.1 & 100    & 1994 & 74.8 & 7.5
  \\ Adam  & 1e-3 & 0.1 & 100    & 1995 & 75.7 & 7.5

\\ \hline 
  \rowcolor{white}
      Mean &      &     &       &      & 75.1  & 7.5
    \\   \rowcolor{white}
    STD  &      &     &       &      & 0.49   & 0.06
\\
     \bottomrule
    \end{tabularx}
    \label{tab:hyper_CUB} 
\end{table*}

\begin{table*}[!htb]
 \rowcolors{6}{gray!10}{gray!40}
  \centering 
  \caption{Impact of size (the number of kernels) of a pool. Pools are pretrained on Sub-ImageNet. Results are averaged on $3$ different random seeds. } 
     \begin{tabularx}{0.9\textwidth}{@{\extracolsep{\fill}} c c c c c c c @{}} 
    \toprule
      \rowcolor{white}
    Size of a Kernel Pool  &  Benchmark &  Average Accuracy (\%) & Memory  \\ [0.5ex] 
    \hline 
    128 & Split-CIFAR100 & 68.9  & 25.6
    \\
    256 & Split-CIFAR100 & 67.5  & 28.8 
    \\
    512 & Split-CIFAR100 & 72.5  & 34.1
    \\
    1024 & Split-CIFAR100 & 66.2 & 39.1
    \\
    128 & Cub-to-Sketch & 54.9  & 7.1
    \\
    256 & Cub-to-Sketch & 48.7  & 8.2
    \\
    512 & Cub-to-Sketch & 67.3  & 10.2
    \\
    1024 & Cub-to-Sketch & 50.6  &  12.1
    \\

     \bottomrule
    \end{tabularx}
    \label{tab:size_kp} 
\end{table*}

\section{Different Model Architectures}
\label{app:arch}
As presented in Section 5.6 in the main paper, we test another 3 different architectures: Resnet-34~\cite{Resnet}, MobileNet-V2~\cite{mobilenetv2}, VGG-16~\cite{VGG}. In this section, we show the detailed implementation and training setups. 
\\
{\bf Model modification.} For Finetune-VGG16, We replace the last 3 fully connected (FC) layers with a single FC layer to reduce the training workload. For Finetune-MobileNet-V2, we remove the Dropout layer in the classifier as our results show Dropout can hinder its performance.
\\
{\bf Pretraining.} The baselines finetune a corresponding pretrained model (pretrained on ImageNet) for each task separately and the pretrained model is taken from Pytorch model zoo. The NWS-Res34 is pretrained for 160 epochs on ImageNet while NWS-VGG16 and  NWS-MobileNetV2 are pretrained for 160 epochs on Sub-ImageNet. 
\\
{\bf Hyperparameters.} We use Adam optimizer with an initial learning rate of $0.01$, a milestone of $\{50\}$, a batch size of $32$ and epochs of $100$ for all methods to train new models. 







\section{Baseline Implementation Details}
\label{app:baseline}
As mentioned in Section 5.3 in the main paper, following the previous works~\cite{CL_KSM_yang2021,CL_PackNet}, we use the same set of common hyperparameters for comparing methods and our method. More specifically, there are $5$ common hyperparameters: the initial learning rate, batch size, training epochs, attributes of the optimizer and the scheduler. Here we show the model-specific hyperparameters as followed.
\begin{itemize}
\item \textbf{Finetune} No extra model-specific hyperparameters.
\item  \textbf{KSM}~\cite{CL_KSM_yang2021}. We adapt from the official repository~\href{https://github.com/LYang-666/CVPR_2021_KSM}{KSM}. We use default values of the learning rate for the mask (2e-4) and initialization of the mask (1e-2). It saves the statistics of batch normalization layers, soft masks for each task, and a single shared backbone for all tasks.
\item \textbf{PackNet}~\cite{CL_PackNet}. We adapt from the repository
\href{https://github.com/ivclab/CPG}{CPG}. The prune ratio is set to the default value of $60\%$. The model is finetuned from the backbone model for $100$ epochs (in default) before pruning is conducted. Then retraining the model for $30$ (in default) epochs. It learns a binary kernel-wise mask for each task and updates the pretrained model. After the first task is learned, it frees up the weights at a predefined ratio (60\%) and relearns the first task to obtain two groups of weights, namely, 60\% learnable weights and 40\% fixed weights. It finetunes the released learnable weights for the second task. After the second task is trained, 60\% of the learnable weights are freed up, which means only 36\% of the total weights remain learnable, and so on. For each task, it saves the statistics of batch normalization layers. Note that all tasks share the same set of kernel masks; so the memory cost of saving kernel masks does not change. The backbone model is updated during the incremental training and only the last updated backbone is saved. The learning of the current model relies on the previous model only.
\item  \textbf{AQD}~\cite{MC_AQD_liu2020}.  We adapt from the official repository~\href{https://github.com/aim-uofa/model-quantization}{AQD}. The bitwidths of feature quantization and weight quantization are both $9$ bits. It saves a quantised network for each task.

\end{itemize}

\section{Visualization of Distributions of Kernel Indices}
\label{app:visual_indices}
To recap, the ResNet-18 has a total of 21 layers (including short-cut layers). Consequently, the layer-wise kernel pool for $8_{th}$, $13_{th}$, $18_{th}$ layers are convolution layers with a kernel size of $1 \times 1$ for short-cut function. The last layer (the $21_{th}$ layer), the replacement of an FC layer, is implemented as $1 \times 1$ convolution layer as well. The layer $1$ has a kernel size of $7 \times 7$ and that of the remaining ones is $3 \times 3$. 

\begin{figure*}[htb]
\begin{tabular}{c}
        \includegraphics[width=0.95\textwidth]{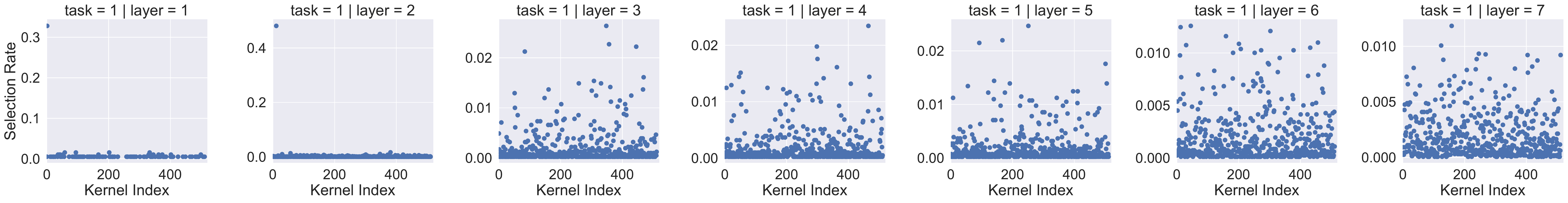} \\ 

        \includegraphics[width=0.95\textwidth]{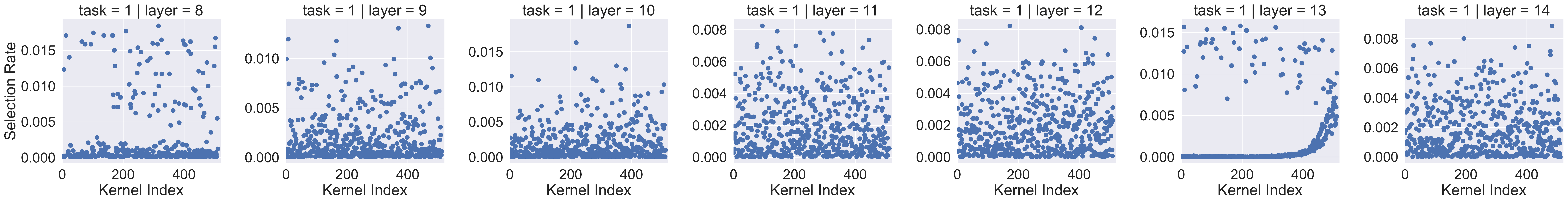}  \\
      \includegraphics[width=0.95\textwidth]{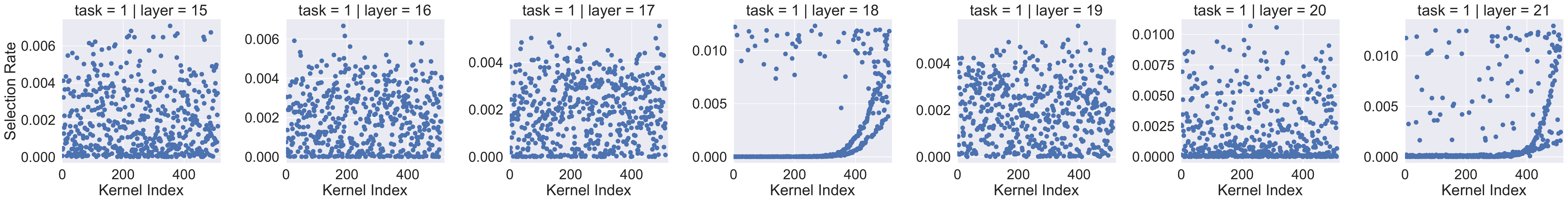}  

\end{tabular} 
\caption{Visualization of the distribution of selected kernel indices of the model trained on CUB~\cite{DS_CUB200}}
\label{fig:vis_cub}
\end{figure*}

\begin{figure*}[htb]

\begin{tabular}{c}
        \includegraphics[width=0.95\textwidth]{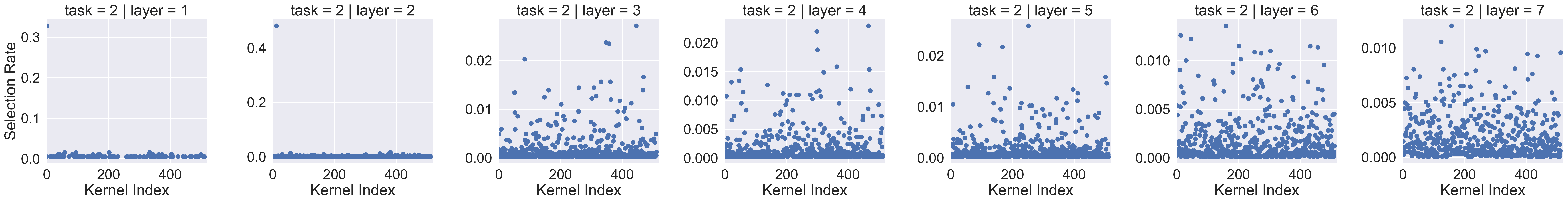} \\ 

        \includegraphics[width=0.95\textwidth]{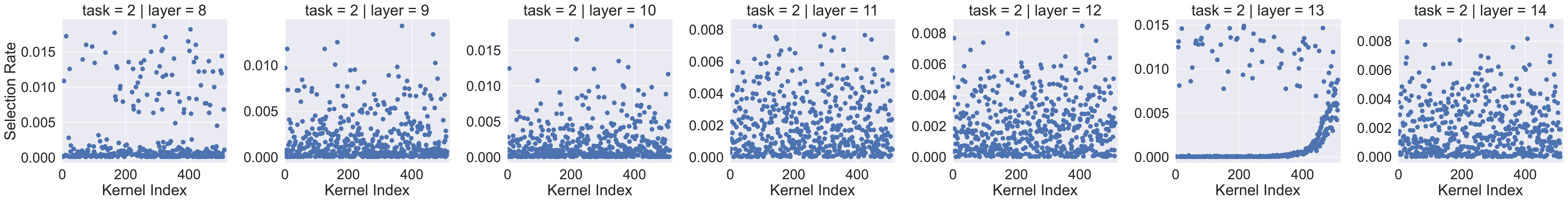}  \\
      \includegraphics[width=0.95\textwidth]{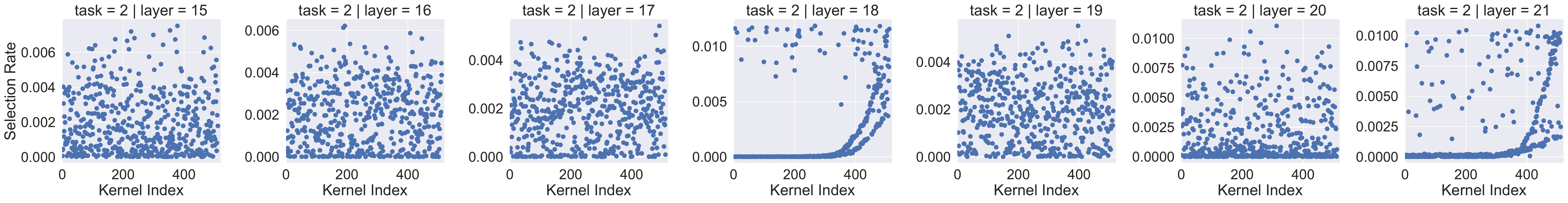}  

\end{tabular} 
\caption{Visualization of the distribution of selected kernel indices of the model trained on Cars~\cite{DS_STANFORD_CARS196}}
\label{fig:vis_cars}
\end{figure*}

\begin{figure*}[htb]

\begin{tabular}{c}
        \includegraphics[width=0.95\textwidth]{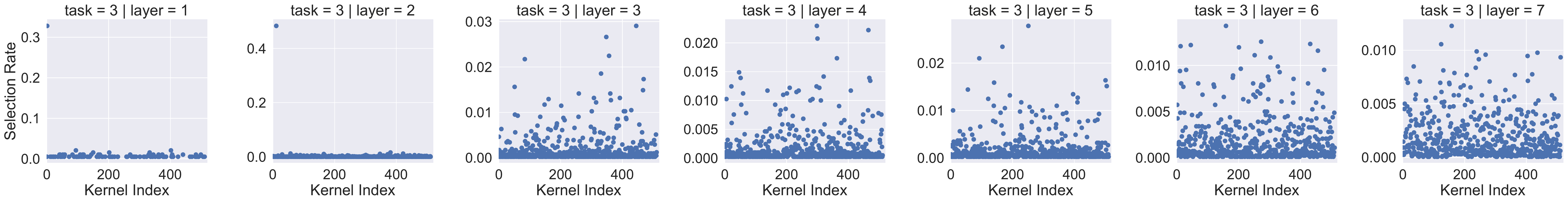} \\ 

        \includegraphics[width=0.95\textwidth]{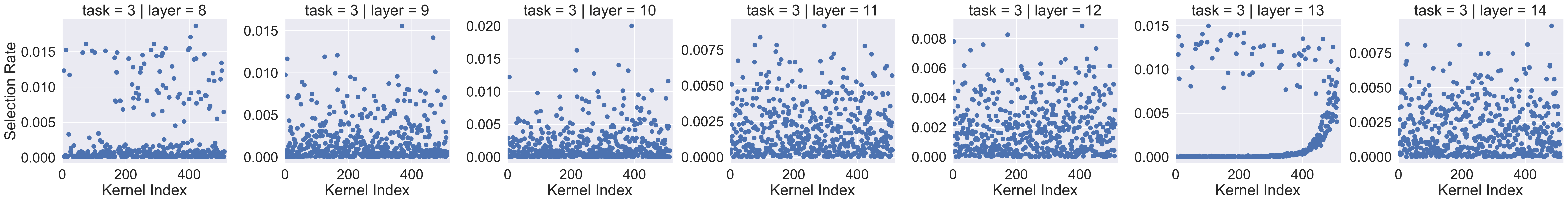}  \\
      \includegraphics[width=0.95\textwidth]{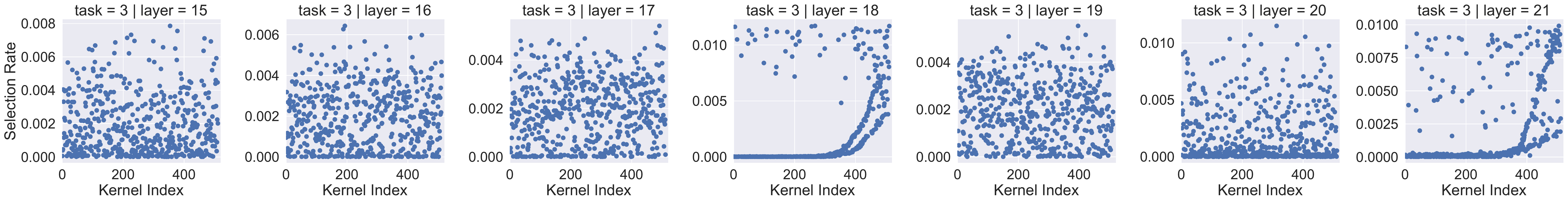}  

\end{tabular} 
\caption{Visualization of the distribution of selected kernel indices of the model trained on Flowers~\cite{DS_OXFORD_FLOWERS102}}
\label{fig:vis_flowers}
\end{figure*}

\begin{figure*}[htb]

\begin{tabular}{c}
        \includegraphics[width=0.95\textwidth]{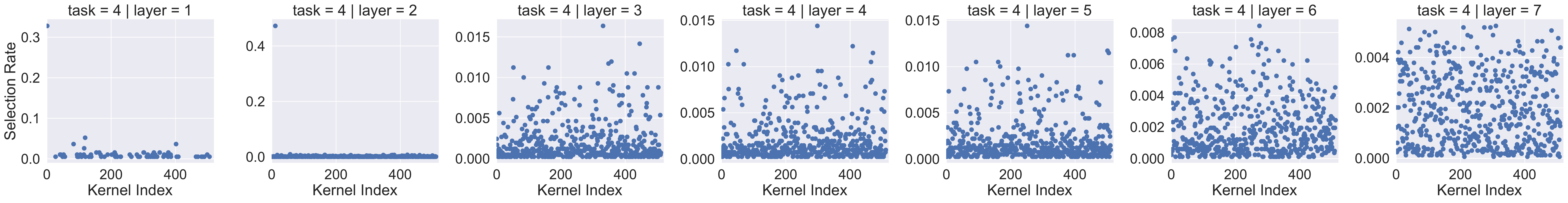} \\ 

        \includegraphics[width=0.95\textwidth]{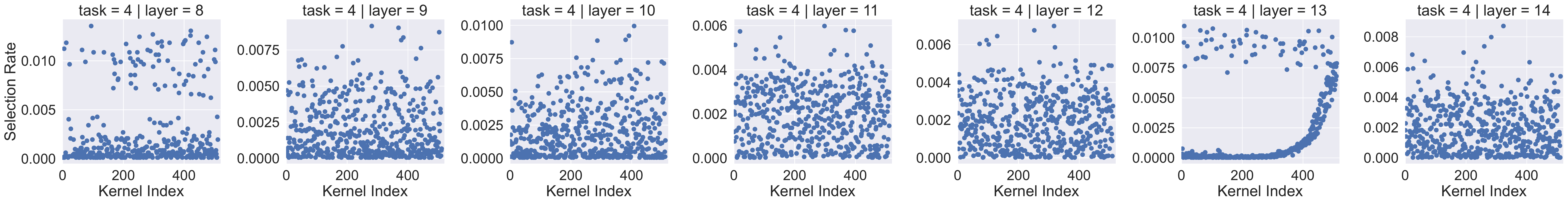}  \\
      \includegraphics[width=0.95\textwidth]{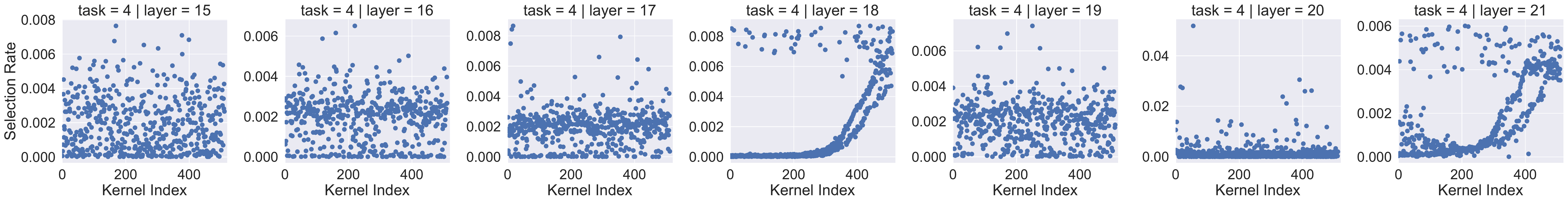}  

\end{tabular} 
\caption{Visualization of the distribution of selected kernel indices of the model trained on WikiArt~\cite{DS_WIKIART195}}
\label{fig:vis_wikiart}
\end{figure*}

\begin{figure*}[htb]
\begin{tabular}{c}
        \includegraphics[width=0.95\textwidth]{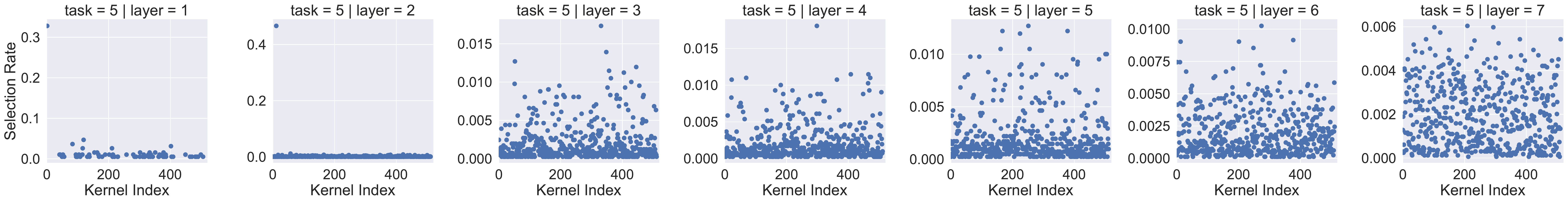} \\ 

        \includegraphics[width=0.95\textwidth]{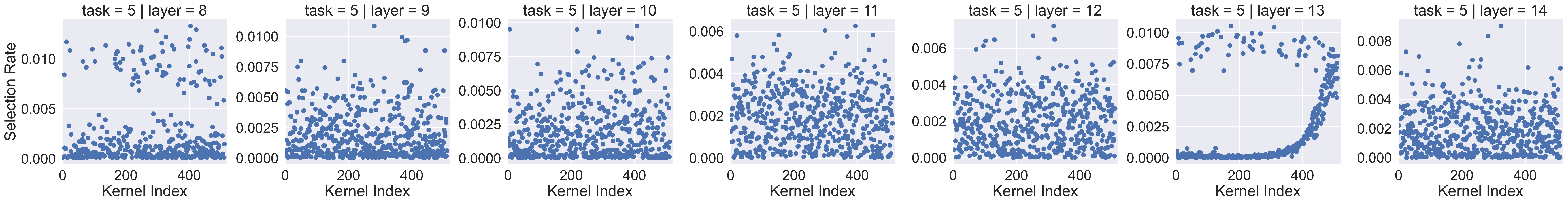}  \\
      \includegraphics[width=0.95\textwidth]{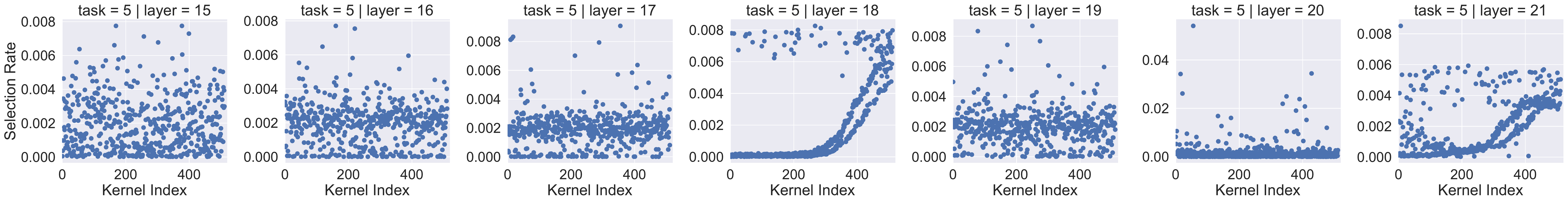}  

\end{tabular} 
\caption{Visualization of the distribution of selected kernel indices of the model trained on Sketches~\cite{DS_SKETCH250}}
\label{fig:vis_sketches}
\end{figure*}

We visualise layer-wise distributions of selected kernel indices of $5$ trained models for CUB-to-Food , for the $21$ layers, from Figure~\ref{fig:vis_cub} to~\ref{fig:vis_sketches} respectively. In each subplot, the x-axis is the kernel index and the y-axis is the Selection Rate (SR) of a unique kernel. We define SR as $h_{i}^{l} / d^{l}$, a frequency of a kernel index $i$ is selected. To recall, $h_{u}^{l}$ is the selection times of a unique index and ${d^{l}}$ is the number of required kernels to build the layer. Note that, SR is in the range of [0, 1] and $i$ is in the range of [0, 511]. We observe that although the distributions of the same layer of the two different tasks are similar, the ordering of indices is different. 

The first two layers among tasks select some kernels with a rate larger than $0.3$, resulting in low diversity of kernels while in other layers the selections of kernels are more diverse. Our observations show that, for the first and the second layers of the first task, more than $30\%$ of the selected kernels are the same kernel. We observe similar distributions of the same layer among different tasks. Intuitively, frequently selected kernels are helpful because of their generality but it is hard to infer the actual factor inducing the dominance. To give some insights, low-level kernels capture coarse common local features (e.g., line, curve, and dot), high-level kernels extract fine-grained specialized global features (e.g., ear, eye, and head), and the last layer infers the class based on high-level features. Therefore, we conjecture that low-level kernels are less diverse and high diversity is necessary for high-level kernels to ensure specialisation. Besides, it may be because the former layers have a less number of kernels, while the later layers are over-parameterized. In other words, more kernels in a layer naturally lead to a more diverse selection of kernels from the corresponding kernel pool. 

\end{appendices}

{\small
\bibliographystyle{ieee_fullname}
\bibliography{egbib}
}

\end{document}